# Learning Task Grouping and Overlap in Multi-Task Learning

Abhishek Kumar                                                                          ABHISHEK@CS.UMD.EDU
Hal Daumé III                                                                           HAL@UMIACS.UMD.EDU
Department of Computer Science, University of Maryland, College Park, MD 20742, USA

## Abstract

In the paradigm of multi-task learning, multiple *related* prediction tasks are learned jointly, sharing information across the tasks. We propose a framework for multi-task learning that enables one to *selectively share* the information across the tasks. We assume that each task parameter vector is a linear combination of a finite number of underlying basis tasks. The coefficients of the linear combination are sparse in nature and the overlap in the sparsity patterns of two tasks controls the amount of sharing across these. Our model is based on the assumption that task parameters within a group lie in a low dimensional subspace but allows the tasks in different groups to overlap with each other in one or more bases. Experimental results on four datasets show that our approach outperforms competing methods.

## 1. Introduction

Multi-task learning is concerned with simultaneously learning multiple prediction tasks that are related to one another (Caruana, 1997; Thrun & Pratt, 1998). The hope is that common information relevant to prediction can be shared among these tasks and learning them jointly can result in better generalization performance than independently learning each task. The key aspect in all multi-task learning methods is the introduction of an inductive bias in the joint hypothesis space of all tasks that reflects our prior beliefs about task relatedness structure. Assumptions that task parameters lie close to each other in some geometric sense (Evgeniou & Pontil, 2004), or parameters share a common prior (Yu et al., 2005; Lee et al., 2007; Daumé III, 2009), or they lie in a low dimensional subspace (Argyriou et al., 2008a) or on a manifold (Agarwal et al., 2010) are some examples of introducing an inductive bias in the hope of achieving better generalization. A major challenge in multi-task learning is how to *selectively* screen the sharing of information so that unrelated tasks do not end up influencing each other. Sharing information between two unrelated tasks can worsen the performance of both tasks. This phenomenon is also known as *negative transfer*. In this paper, we propose a structured prior on the tasks' weight matrix (whose columns are parameter vectors for individual prediction tasks) that allows formation of groups of related tasks with partial overlap among the groups. Two tasks can have full, partial or no overlap, which is determined by number of *basis tasks* they share. We use the term *task* to refer to the actual prediction problem. We will use the term *task weight or parameter vector* to refer to the parameters of the model learned from training data.

A starting point for our method is the assumption that task parameters lie in a low dimensional subspace (Argyriou et al., 2008a). This is achieved by imposing a trace-norm constraint on the task weight matrix. However, the low rank assumption does not differentiate between the tasks and assumes that all tasks are related, which may adversely affect the performance when there are unrelated tasks in the pool, or some tasks have more in common than others. One way to address this problem is to assume that there are disjoint groups of tasks. Examples of such approaches are (Jacob et al., 2008; Xue et al., 2007) where tasks are assumed to be clustered and parameters of tasks within a cluster lie close to each other in $\ell_2$ norm sense. Our proposed method does not regularize based on the $\ell_2$ distance between task parameters which can fail to take advantage of negative correlation between the tasks. For example, grouping on the basis of $\ell_2$ distance can put two tasks with parameters $\mathbf{w}$ and $-\mathbf{w}$ in separate clusters and block the sharing of information between them while there is clearly a relation between the tasks, i.e., the span of these two parameter vectors is one dimensional sub-space instead of being a two dimensional space. An inductive bias that regu-larizes based on the subspace assumption could have exploited the task relatedness of this sort. We do not assume disjoint groups and allow partial overlap between them.

Recently, task grouping in the subspace based regularization framework was proposed in (Kang et al., 2011). Tasks are assumed to form disjoint groups and the tasks within each group are assumed to lie in a low dimensional subspace. The parameters of tasks and the group assignment matrix are both learned using alternating style optimization that converges to a local minimum. However, the subspaces shared by each group do not have any overlap between them, which may not always reflect the true sharing structure since there is often a continuum in the sharing between tasks. One pair of tasks may have more in common than another task pair and we may not be able to take full advantage of multi-task learning by putting the two tasks in the second pair in different groups.

In our model, we assume that task parameters within a group lie in a low dimensional subspace, and allow two tasks from different groups to overlap by having one or more bases in common. This is achieved by assuming that there exist a small number of latent basis tasks and parameter vector of every observed task is a linear combination of these. If the columns of $\mathbf{L}$ denote the parameter vectors of $k$ latent tasks, we model the parameter vector $\mathbf{w}_t$ of observed task $t$ as $\mathbf{w}_t = \mathbf{L}\mathbf{s}_t$, where $\mathbf{s}_t$ contain the coefficients of the linear combination. However, each linear combination is assumed to be sparse in the latent bases and the overlap in the sparsity patterns of any two tasks controls the amount of sharing between these. The low dimensional subspace assumption of (Argyriou et al., 2008a) can also be thought of having a small number of latent basis tasks, however each observed task is determined by a full linear combination of these bases and there is no notion of task groups. To be more clear about the difference, if there are $k$ latent tasks and we pick $r(\leq k)$ observed tasks arbitrarily, the corresponding weight matrix of these tasks will be of rank $r$ in the model of (Argyriou et al., 2008a). On the other hand, our model allows this matrix to be of rank less than $r$ by imposing sparse structure on the linear combination weights. We validate our approach empirically with two synthetic and four real-world datasets and observe that our method either outperforms or performs as well as the relevant baseline methods of (Kang et al., 2011; Argyriou et al., 2008a).

## 2. Related Work

Several methods have been proposed in the literature for the problem of multi-task learning. Most methods work on the assumption that all tasks are related (Evgeniou & Pontil, 2004; Ando & Zhang, 2005; Argyriou et al., 2008a; Rai & Daumé III, 2010). This assumption can be violated in many real applications and can degrade the performance. To avoid this, several methods have been proposed to allow for grouping of the tasks using different notions of grouping. Some methods assume that tasks can be grouped in clusters and parameters of tasks within a cluster are either close to each other in some distance metric or share a common probabilistic prior (Bakker & Heskes, 2003; Jacob et al., 2008; Xue et al., 2007; Zhou et al., 2011). Tasks in different clusters do not interact with one another. However, these methods might fail to take advantage of negatively correlated tasks since they can put these in different clusters. A similar idea was used in (Thrun & O'Sullivan), where two tasks are taken to be similar if one's parameters improve performance on the other task.

Other methods assume that there is one group of related tasks and a small number of outlier tasks that are not related to any task in the pool (Yu et al., 2007; Chen et al., 2011). There also exist probabilistic models which attempt to learn full task covariance matrix and use it in learning of predictor functions (Zhang & Yeung, 2010b;a; Zhang & Schneider, 2010; Archambeau et al., 2011). These methods place a matrix variate prior on the task matrix $\mathbf{W}$.

Another common assumption is that task parameters lie in a low dimensional subspace that captures the predictive structure for all the tasks (Argyriou et al., 2008a; Liu et al., 2009). These methods assume that some of features (either in original space, or in a transformed space) are inactive for all tasks. This forces all task parameters to lie in a low dimensional subspace. In (Jalali et al., 2010), this model was refined and features are assumed to be either active for all tasks, or inactive for *most* of the tasks. This is done by forcing $\mathbf{W}$ to be sum of a group sparse matrix and a sparse matrix, hence predictors no longer lie in a low dimensional subspace.

There exist a few methods that incorporate grouping structure in the subspace based regularization (Argyriou et al., 2008b; Kang et al., 2011). In (Argyriou et al., 2008b), tasks in each group share a common linear transformation for feature extraction. It is shown to be equivalent to minimizing the trace-norm of each groups' weight sub-matrix. The objective is non-convex and it is optimized using stochastic gradient descent. Very similar in spirit is the work of (Kang et al., 2011), where tasks within a group are assumed to lie in a low dimensional subspace and they minimize the square of trace-norm of each group's weight sub-matrix (instead of trace-norm as in (Argyriou et al., 2008b)). The non-convex objective is optimized using mixed integer programming. Both these methods assume that groups are disjoint and tasks are either related (within a group) or totally unrelated (in different groups). This is in contrast to the approach proposed in this paper, where the low dimensional subspace shared by group members is not exclusive to it, and two tasks from different groups are allowed to overlap in one or more bases. Intuitively, this means that some of the latent basis tasks influence more than one group. Recently, (Passos et al., 2012) posited a mixture of sparse factor analyzers structure on the collection of the task weight vectors. Their model assumes that the tasks form clusters and tasks within each cluster are a *sparse* combination of a task dictionary specific to that cluster. A nonparametric Bayesian approach is used to learn the number of clusters and the task dictionary sizes from the data. The task clusters do not have overlap in task dictionary elements.

## 3. Learning Task Grouping and Overlap

In this Section, we describe our approach for modeling task grouping and overlap. We call the proposed approach as GO-MTL for Grouping and Overlap in Multi-Task Learning. Suppose we have $T$ tasks and $Z_t = \{(\mathbf{x}_{ti}, y_{ti}) : i = 1, 2, \ldots, N_t\}$ be the training set for each task $t = 1, 2, \ldots, T$. Let $\mathbf{w}_t$ represent the weight vector for task indexed by $t$. These task weight vectors are stacked as columns of a matrix $\mathbf{W}$, which is of size $d \times T$, with $d$ being the feature dimension.

We assume there are $k(< T)$ latent basis tasks and each observed task can be represented as linear combination of a *subset* of these basis tasks. This assumption enables us to write the weight matrix $\mathbf{W}$ as $\mathbf{W} = \mathbf{LS}$, where $\mathbf{L}$ is a matrix of size $d \times k$ with each column representing a latent task, and $\mathbf{S}$ is a matrix of size $k \times T$ containing the weights of linear combination for each task. The predictor $\mathbf{w}_t$ for task $t$ is given by $\mathbf{Ls}_t$, where $\mathbf{s}_t$ is $t$'th column of matrix $\mathbf{S}$. We assume the matrix $\mathbf{S}$ to be sparse to enforce that each observed task is obtained from only a few of the latent tasks, indexed by the non-zero pattern of the corresponding column of matrix $\mathbf{S}$. The predictive structure of the tasks is captured by the matrix $\mathbf{L}$ and the grouping structure is determined by matrix $\mathbf{S}$. For two columns $\mathbf{s}_{t_1}$ and $\mathbf{s}_{t_2}$ of matrix $\mathbf{S}$ corresponding to tasks $t_1$ and $t_2$, the overlap between the sparsity patterns determines the number of basis tasks they have in common. Tasks that have same sparsity pattern can be seen as belonging to same *group*, while tasks whose sparsity patterns are orthogonal to each other can be seen as belonging to different groups. The partial sharing of bases allows us to do away with the concept of disjoint groups, and allows to model tasks which are not as much related as they are with tasks in their own group but which still have something in common. The task that does not share bases with any other task in the pool can be seen as *outlier task*.

If $\mathbf{s}_t$ denotes the sparsity pattern for task $t$, our learning cost function takes the following form:

$$\sum_t \sum_{(\mathbf{x}_{ti}, y_{ti}) \in Z_t} \mathcal{L}(y_{ti}, \mathbf{s}_t' \mathbf{L}' \mathbf{x}_{ti}) + \mu ||\mathbf{S}||_1 + \lambda ||\mathbf{L}||_F^2, \quad (1)$$

where $\mathcal{L}(\cdot, \cdot)$ is the empirical loss function, $|| \cdot ||_1$ is entry-wise $\ell_1$ norm of the matrix and $||\mathbf{L}||_F = (tr(\mathbf{LL}'))^{1/2}$ is the Frobenius norm of matrix $\mathbf{L}$. The parameter $\mu$ controls the sparsity in $\mathbf{S}$. The penalty on the Frobenius norm of $\mathbf{L}$ regularizes the predictor weights to have low $\ell_2$ norm and avoids overfitting.

For a convex empirical loss function, the cost function in Eq. 1 is convex in $\mathbf{L}$ for a fixed $\mathbf{S}$, and is convex in $\mathbf{S}$ for a fixed $\mathbf{L}$, however, it is not jointly convex. We adopt alternating optimization strategy that converges to a local minimum. For a fixed $\mathbf{L}$, the optimization function can be decomposed in individual problems for $\mathbf{s}_t$ as

$$\mathbf{s}_t = \arg\min_{\mathbf{s}} \sum_{(\mathbf{x}_{ti}, y_{ti}) \in Z_t} \mathcal{L}(y_{ti}, \mathbf{s}' \mathbf{L}' \mathbf{x}_{ti}) + \mu ||\mathbf{s}||_1, \quad (2)$$

We use two-metric projection method to optimize Eq. 2, which has superlinear convergence (Schmidt et al., 2007; Gafni & Bertsekas, 1984). For a fixed $\mathbf{S}$, the optimization problem reduces to following:

$$\min_{\mathbf{L}} \sum_{t=1}^{T} \sum_{(\mathbf{x}_{ti}, y_{ti}) \in Z_t} \mathcal{L}(y_{ti}, \mathbf{s}_t' \mathbf{L}' \mathbf{x}_{ti}) + \lambda ||\mathbf{L}||_F^2. \quad (3)$$

This problem is convex in $\mathbf{L}$ and has a closed form solution for squared loss function, which is commonly used in regression problems. For classification problems, we use logistic loss and optimize it using Newton-Raphson method, which is commonly used to estimate logistic regression parameters and is the basis of iterative reweighted least squares algorithm (IRLS) algorithm for logistic regression (Green, 1984). We also experimented with steepest gradient descent and found it to work reasonable well on all datasets that we tried.

Algorithm 1 outlines the steps and initialization procedure for our approach. We adopt the following strategy for initializing $\mathbf{L}$. The individual task parameters

**Algorithm 1** GO-MTL: Grouping and Overlap for Multi-Task Learning

**Input:**
$Z^t$: Labeled training data for all tasks
$k$: Number of latent tasks
$\mu$: Parameter for controlling sparsity
**Output:** Task predictor matrix $\mathbf{W}$, $\mathbf{L}$ and $\mathbf{S}$.
1: Learn individual predictors for each task using only its own data.
2: Let $\mathbf{W}^0$ be the matrix that contains these initial predictors as columns.
3: Compute top-$k$ singular vectors: $\mathbf{W}^0 = \mathbf{U}\mathbf{\Sigma}\mathbf{V}^T$
4: Initialize $\mathbf{L}$ to first $k$ columns of $\mathbf{U}$.
**while** *not converged* **do**
  **for** $t = 1$ to $T$ **do**
    5: Solve Eq. 2 to obtain $\mathbf{s}_t$.
  **end for**
  6: Construct matrix $\mathbf{S} = [\mathbf{s}_1 \mathbf{s}_2 \ldots \mathbf{s}_T]$.
  7: Save the previous $\mathbf{L}$: $\mathbf{L}_{\text{old}} = \mathbf{L}$.
  8: Fix $\mathbf{S}$ and solve Eq. 3 to obtain $\mathbf{L}$.
**end while**
9: Return outputs: $\mathbf{L} = \mathbf{L}_{\text{old}}$, $\mathbf{S}$ and $\mathbf{W} = \mathbf{L}_{\text{old}}\mathbf{S}$.

are learned independently using their own data without any sharing, which are then stacked as columns in a weight matrix $\mathbf{W}^0$. The matrix $\mathbf{L}$ is then initialized to the top-$k$ left singular vectors of $\mathbf{W}^0$. These are the directions that capture maximum variance of task parameters in a $k$-dimensional space. This initialization strategy was observed to be effective in all our experiments. The alternating optimization procedure is terminated when there is little change in $\mathbf{L}$ or $\mathbf{S}$ between two consecutive iterations.

The parameter $k$ determines the number of latent tasks, which is taken to be less than total number of tasks $T$. The amount of inductive bias depends on this number. In this respect, it is similar to the "number of groups" parameter, $G$, in (Kang et al., 2011). If $k$ is very low, it may shrink the hypothesis space too much. On the other hand, if $k$ is very high, the tasks are not forced to share information with each other. When we increased the value of $k$ in the experiments starting from 1, the prediction accuracy improves in the beginning. After a certain value of $k$, the performance becomes stable and the possible decrease in performance due to large $k$ can be controlled by increasing the sparsity penalty $\mu$. More details on this behavior are provided in Sec. 4.

It is possible to have an alternative formulation to Eq. 1 where we can do away with parameter $k$ (i.e., make it equal to $T$), and instead enforce a low rank penalty on matrix $\mathbf{L}$, weighted by a parameter $\alpha$. This can be done by penalizing the nuclear norm of $\mathbf{L}$, which is the tightest convex lower bound on the rank function in the unit ball of matrices (i.e., matrices with spectral norm less than one). However, there are two disadvantages to this approach: (a) this convex surrogate is not always guaranteed to produce a low rank solution, and (b) this will result in a non-smooth optimization problem for $\mathbf{L}$ due to non-smoothness of the nuclear norm.

### 3.1. Regression: Squared Loss

Here, we give details about optimization of the cost function of Eq. 1 for squared loss $\mathcal{L}(a,b) = (a-b)^2$, commonly used in regression problems. Let us denote $\mathbf{y}_t$ to be a column vector of length $N_t$ that contains all the labels for task $t$. Similarly, let $\mathbf{X}_t$ be the data matrix of size $d \times N_t$ containing all the samples for task $t$ stacked as columns. The cost function of Eq.1 can be written as,

$$\min_{\mathbf{L},\mathbf{S}} \sum_{t=1}^{T} \frac{1}{N_t} ||\mathbf{y}_t - \mathbf{X}_t'\mathbf{L}\mathbf{s}_t||^2 + \mu||\mathbf{S}||_1 + \lambda||\mathbf{L}||_F^2, \quad (4)$$

For a fixed $\mathbf{L}$, we need the gradient and Hessian of the squared loss function $f(\mathbf{s}_t) = \frac{1}{N_t}||\mathbf{y}_t - \mathbf{X}_t'\mathbf{L}\mathbf{s}_t||^2$ to optimize for $\mathbf{s}_t$ using two-metric projection method. These are given as $\nabla_{\mathbf{s}_t} f(\mathbf{s}_t) = \frac{2}{N_t}\mathbf{L}'\mathbf{X}_t(\mathbf{X}_t'\mathbf{L}\mathbf{s}_t - \mathbf{y}_t)$, and $\nabla_{\mathbf{s}_t}^2 f(\mathbf{s}_t) = \frac{2}{N_t}\mathbf{L}'\mathbf{X}_t\mathbf{X}_t'\mathbf{L}$. For a fixed $\mathbf{S}$, equating the gradient of Eq. 4 to zero gives

$$\sum_{t=1}^{T} \frac{1}{N_t}\mathbf{X}_t\mathbf{y}_t\mathbf{s}_t' = \sum_{t=1}^{T} \frac{1}{N_t}\mathbf{X}_t\mathbf{X}_t'\mathbf{L}\mathbf{s}_t\mathbf{s}_t' + \lambda\mathbf{L}$$

This is a linear equation in $\mathbf{L}$. To solve this, we apply vectorization operator on both sides, which simply stacks all columns of a matrix one above another and forms a long vector. Clearly, this is a linear operator and can pass through the summation, and we obtain

$$\sum_{t=1}^{T} \frac{1}{N_t}\text{vec}\left(\mathbf{X}_t\mathbf{y}_t\mathbf{s}_t'\right) = \text{vec}\left(\sum_{t=1}^{T} \frac{1}{N_t}\mathbf{X}_t\mathbf{X}_t'\mathbf{L}\mathbf{s}_t\mathbf{s}_t' + \lambda\mathbf{L}\right)$$

$$= \left[\sum_{t=1}^{T} \frac{1}{N_t}(\mathbf{s}_t\mathbf{s}_t') \otimes (\mathbf{X}_t\mathbf{X}_t') + \lambda\mathbf{I}\right]\text{vec}(\mathbf{L}),$$

where we have used a property of Kronecker product that $\text{vec}(\mathbf{A}\mathbf{X}\mathbf{B}) = (\mathbf{B}' \otimes \mathbf{A})\text{vec}(\mathbf{X})$. This is in standard form of system of linear equations that is full rank and has a unique solution. It can be solved using LU decomposition or by iterative methods, which are much faster and numerically more stable then solving it using matrix inverse.

### 3.2. Classification: Logistic Loss

We use logistic regression for classification problems, although the proposed method is not tied to any par-ticular loss function. Here, we give details about optimizing Eq. 1 for logistic loss function, which is given as $\mathcal{L}(y, f(\mathbf{x})) = \log(1 + \exp(-y f(\mathbf{x})))$, where $y \in \{-1, 1\}$ is the true label. Let us denote the logistic function by $\sigma(x) = 1/(1 + e^{-x})$. For a fixed $\mathbf{L}$, we need the gradient and Hessian of the loss function w.r.t. $\mathbf{s}_t$ to solve using two-metric projection method, which are given by

$$f(\mathbf{s}_t) = \frac{1}{N_t} \sum_{i=1}^{N_t} \log(1 + \exp\left(-y_{ti} \mathbf{s}_t' \mathbf{L}' \mathbf{x}_{ti}\right))$$

$$\nabla_{\mathbf{s}_t} f(\mathbf{s}_t) = \frac{-1}{N_t} \sum_{i=1}^{N_t} (y_{ti} - \sigma(\mathbf{w}_t' \mathbf{x}_{ti})) \, \mathbf{L}' \mathbf{x}_{ti}$$

$$\nabla^2_{\mathbf{s}_t} f(\mathbf{s}_t) = \frac{1}{N_t} \sum_{i=1}^{N_t} \sigma(\mathbf{w}_t' \mathbf{x}_{ti})(1 - \sigma(\mathbf{w}_t' \mathbf{x}_{ti})) \mathbf{L}' \mathbf{x}_{ti} \mathbf{x}_{ti}' \mathbf{L}$$

where $\mathbf{w}_t = \mathbf{L}\mathbf{s}_t$ is the weight vector for task $t$. For a fixed $\mathbf{S}$, the objective is again convex in $\mathbf{L}$ and we give both gradient update and Newton-Raphson update here.

$$\nabla_{\mathbf{L}} : -\sum_{t=1}^{T} \frac{1}{N_t} \sum_{i=1}^{N_t} (y_{ti} - \sigma(\mathbf{w}_t' \mathbf{x}_{ti})) \, \mathbf{x}_{ti} \mathbf{s}_t' + 2\lambda \mathbf{L}$$

For Newton-Raphson update, we use Taylor series expansion up to second order around $\mathbf{L}$, making use of directional first and second derivatives. The step direction $\mathbf{M}$ is obtained by solving the following system of linear equations:

$$\left[\!\!\left[\sum_{t=1}^{T} \frac{1}{N_t} \sum_{i=1}^{N_t} \delta_{ti} \left[\text{vec}(\mathbf{x}_{ti}\mathbf{s}_t') \otimes \text{vec}(\mathbf{x}_{ti}\mathbf{s}_t')'\right] + 2\lambda \mathbf{I}\right]\!\!\right] \text{vec}(\mathbf{M})$$
$$= \text{vec}\left(\sum_{t=1}^{T} \frac{1}{N_t} \sum_{i=1}^{N_t} (y_{ti} - \sigma(\mathbf{w}_t' \mathbf{x}_{ti})) \, \mathbf{x}_{ti}\mathbf{s}_t' - 2\lambda \mathbf{L}\right)$$

where $\delta_{ti} = \sigma(\mathbf{w}_t' \mathbf{x}_{ti})(1 - \sigma(\mathbf{w}_t' \mathbf{x}_{ti}))$. The Newton-Raphson update is then carried out by taking a step in this direction. The step size is computed using Armijo rule.

Newton-Raphson updates, although more costly to compute, can converge in a smaller number of iterations. Gradient updates also seemed to work reasonably well in the experiments. This can be considerably faster than Newton-Raphson, especially for large problems, since Newton-Raphson involves solving a system of linear equations of size $dk$ multiple times for every iteration of $\mathbf{L}$.

## 4. Experiments

We perform extensive empirical evaluation of our approach to gauge its effectiveness. We carry out empir-

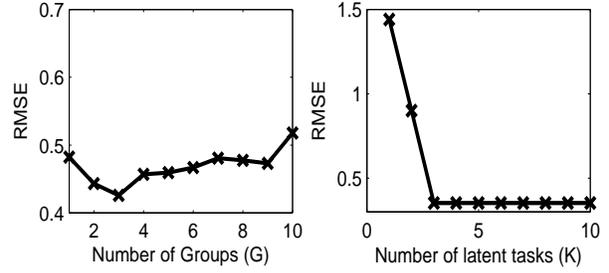

Figure 1. Results on first Synthetic data with disjoint groups. *Left:* RMSE with DG-MTL (Kang et al., 2011) vs. number of groups $(G)$, *Right:* RMSE with GO-MTL vs. number of latent tasks $(k)$

ical comparisons with following two competing subspace regularized multi-task learning approaches:

- **No-group MTL** (Argyriou et al., 2008a): All tasks are assumed to be related, and the task parameters are assumed to lie in a low dimensional subspace. This is done by penalizing the nuclear norm of weight matrix.
- **Disjoint-Group MTL (DG-MTL)** (Kang et al., 2011): A recently proposed approach that assumes multiple disjoint groups of tasks. Task parameters within a group lie in a low dimensional space.

In addition, we also compare with baseline **single task learning (STL)**, in which tasks are learned independently. Below, we report results on two synthetic and four real-world datasets. The regularization parameter $\lambda$ in Eq. 1 is kept fixed at 0.1 in all experiments. We expect $\lambda$ to depend on the dimensionality of the space and number of examples. Incidentally, all the real-world datasets used in this work have dimensionality less than 100, for which $\lambda = 0.1$ seemed to work well. Of course, it can always be selected using cross-validation. We generate four different random splits (70% train, 30% validation) on the training set for cross-validation of parameter $\mu$. The search grid was taken to be $[0.001, 0.005, 0.01, 0.05, 0.1, 0.2, 0.3, 0.4]$. Averaged performance for different splits is reported.

### 4.1. Synthetic data

We use two synthetic datasets to study our approach. First, we use the synthetic data used in (Kang et al., 2011).[1] This data consists of 20-dimensional feature vectors, three groups of tasks, 15 training points and 50 test points per task. There are 10 tasks in each group whose parameter vectors are identical to each other up to a scaling factor. These parameters are used

---

[1] This data along with the source code was taken from author's website: http://www-scf.usc.edu/ zkang/GoupMTLCode.zip

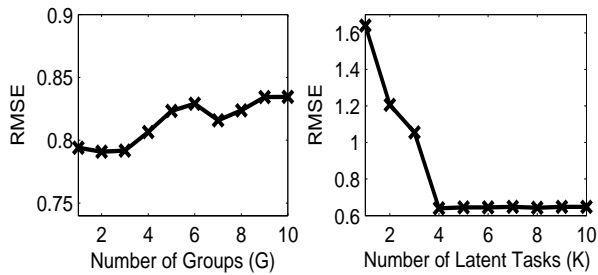

Figure 2. Results on second Synthetic data with overlapping groups. *Left:* RMSE with DG-MTL vs. number of groups ($G$), *Right:* RMSE with GO-MTL vs. number of latent tasks ($k$)

in the model of linear regression to generate training data. The task groups in this data are disjoint.

We generate a second synthetic dataset to simulate overlap in groups. We retain the previous setting of 3 groups and 10 tasks in each group, but now we allow the groups to overlap in one basis. We generate parameter vectors for 4 latent tasks in 20 dimensions, with each entry drawn i.i.d. from a zero mean and unit variance normal distribution. This is essentially the matrix **L** in our formulation. We generate the first 10 tasks by linearly combining only first two latent tasks. The coefficients of linear combination are drawn i.i.d. from a normal distribution centered at zero with unit variance. In a similar manner, we generate the next 10 tasks by linearly combining second and third latent tasks. Last 10 task parameters are generated by linear combination of the last two latent tasks. The matrix **S** in our formulation that contains the coefficients of linear combination, has precisely two non-zero entries in each column for this generative model. We randomly generate 15 training and 50 test points per task, and task parameters are used to generate their real valued labels using a linear regression model. Random Gaussian noise with zero mean and 0.5 standard deviation is added to the labels.

The plot of root mean square error (RMSE) with changing $k$ is shown in Fig. 1 and Fig. 2. We also show the RMSE plot with changing value of parameter $G$ (the number of groups) in the approach of (Kang et al., 2011). GO-MTL converges to almost same RMSE for all values of $k \geq 3$ for first synthetic data and $k \geq 4$ for the second synthetic data. The performance of (Kang et al., 2011) is more sensitive to the number of groups parameter ($G$) and starts deteriorating when it is increased or decreased from the true value. The proposed approach outperforms disjoint-group MTL by a significant margin, more so on the second dataset that has overlap in groups. Ta-ble 1 shows the exact RMSE values obtained for these datasets.

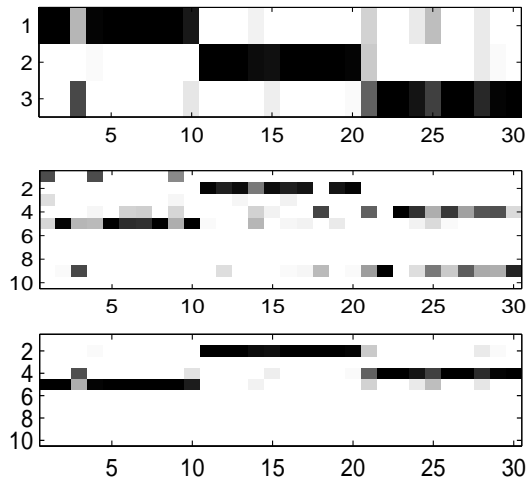

Figure 3. Recovered sparsity patterns (the matrix **S**) with GO-MTL for the first Synthetic data with 3 disjoint groups (darker color indicates higher absolute value of the coefficient). Along horizontal and vertical axes are the observed tasks and the latent tasks, respectively. *Top:* For $k = 3$, *Middle:* For $k = 10$ after three iterations, *Bottom:* For $k = 10$ after convergence (15 iterations). Even for large value of $k$, we are able to recover the true support which is given by three latent tasks.

The sparsity patterns recovered by GO-MTL for these two data are shown in Fig. 3 and Fig. 4. We are able to recover the grouping and overlap structure for most of the tasks in both cases. For the second synthetic data (Fig. 4), support of first 10 and last 10 tasks is recovered more precisely than the support of the middle group, where a few tasks (for example, 11th, 15th and 16th task) have non-negligible coefficients not belonging to the true support. The recovery of support was found to be robust to the value of $k$ chosen in the algorithm, as is shown in the figures. We are able to recover the support with same precision for values of $k \geq 3$ for the first data and values of $k \geq 4$ for the second data.

### 4.2. Real datasets

We evaluate the proposed approach on the following four real-world datasets, two of which are regression tasks and the other two are classification tasks. We treat multi-way classification as multi-task learning problem where each task is the classification of one class from all other classes. To be fair in our comparisons, we evaluate on datasets that are used in (Argyriou et al., 2008a; Kang et al., 2011).

|  | Synth. (1) | Synth. (2) | Computer | School | MNIST | USPS |
|---|---|---|---|---|---|---|
| STL | 1.04 | 1.36 | 2.70 (0.10) | 10.67 (0.20) | 14.8 (0.34) | 9.0 (0.4) |
| No-group MTL | 0.48 | 0.79 | 2.06 (0.07) | 10.18 (0.15) | 14.4 (0.28) | 7.8 (0.2) |
| DG-MTL | 0.42 | 0.80 | 2.01 (0.10) | 10.18 (0.20) | 14.0 (0.30) | 7.8 (0.2) |
| GO-MTL | **0.35** | **0.64** | **1.76** (0.09) | **10.04** (0.24) | **13.4** (0.30) | **7.2** (0.2) |

Table 1. Results on different datasets: Reported numbers are root mean square error (RMSE) for regression datasets and multi-class classification errors for MNIST and USPS. Numbers in parentheses are std. dev., which were negligible for synthetic datasets and so are not reported. STL: Single task learning, No-group MTL (Argyriou et al., 2008a), DG-MTL (Kang et al., 2011), GO-MTL: the proposed method.

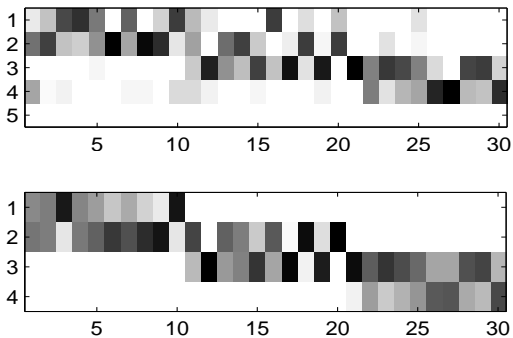

Figure 4. Sparsity patterns (the matrix **S**) for the second Synthetic data with three overlapping groups generated by 4 latent tasks. *Bottom:* True task grouping structure, *Top:* Recovered support for $k = 5$. We can see that fifth latent basis has all coefficients equal to zero and only first four latent tasks are active. The first and third groups are recovered more faithfully than the second group.

- **Computer Survey data:** This regression dataset has been widely in the literature to evaluate various multi-task learning approaches (Argyriou et al., 2008a; Agarwal et al., 2010). The data was collected in a survey of 190 persons who rated their likelihood of purchasing each of 20 different personal computers. Here, students correspond to tasks and computers correspond to examples. All computers are rated by each student on a scale of 0-10, thus giving 20 labeled examples per task. Each computer is represented by 13 different features (RAM, cache-size, hard-disk, CPU speed, etc.). We added one more feature of constant 1 in all examples to account for the bias term in the regression. Training and test set are obtained by splitting the datasets 75%-25%, thus giving 15 examples for training and 5 examples for test.
- **School data:** This regression data has been used in previous works in multi-task learning (Argyriou et al., 2008a; Agarwal et al., 2010; Bakker & Heskes, 2003). This dataset is from Inner London Education Authority and consists of examination scores of 15362 students from 139 schools in London. Here, each school corresponds to a task, thus giving a total of 139 tasks. The input consists of the year of examination, 4 school-specific attributes and 3 student-specific attributes. Following (Argyriou et al., 2008a), each categorical feature is replaced with binary features, giving a total of 26 features. We again add a feature of constant 1 in all examples to account for the bias term. Training and test set are obtained by dividing examples of each task 60%-40%. Number of examples in each task are different; there are about 65 examples per task on average in training and 45 examples per task for test.
- **USPS Digits data:** This is a handwritten digits dataset (Hull, 1994) with 10 classes.[2] The images are processed using PCA and dimensionality is reduced to 87, retaining almost 95% of the variance.
- **MNIST Digits data:** This is another handwritten digits dataset (LeCun et al., 1998) with 10 classes.[2] The images are preprocessed with PCA and dimensionality reduced to 64.

For MNIST and USPS datasets, we use the same setup as in (Kang et al., 2011) where 1000, 500 and 500 samples are used for training, validation and test respectively. The results are summarized in Table 1. All multi-task learning approaches are able to outperform single task learning, however on School data, the improvement is not statistically significant. The proposed method is able to outperform both no-group MTL and disjoint-group MTL.

## 5. Conclusion

We proposed a novel framework for learning grouping and overlap structure in multi-task learning, where parameters of each task group are assumed to lie in a low dimensional subspace. Our approach does not assume disjoint grouping structure, and tasks belonging to different groups are allowed to overlap with each other through sharing of one or more latent basis tasks. This is a more realistic assumption since we can have tasks in our pool that are not related enough to be in the same group, but still share some information that can be exploited for better learning. We validated our model on two synthetic and four real datasets, and obtained considerable gains compared to other competing approaches for subspace regularized multi-task learning that either do not take grouping structure into account, or assume that tasks in different groups do not interact at all. For future work, we would like to extend the proposed model to learn other types of structured interaction patterns among the tasks, e.g., hierarchies of tasks.

---

[2] We thank the authors of (Kang et al., 2011) for providing us the data used in their paper.